\documentclass[sigconf]{acmart}
\usepackage{amsmath,epsfig}
\usepackage{amsmath,amsfonts}
\usepackage{algorithmic}
\usepackage{textcomp}
\usepackage{color}
\usepackage{booktabs}
\usepackage{multirow}
\usepackage{array}
\usepackage[switch]{lineno}  
\usepackage{color}
\usepackage{subfigure}
\usepackage{bm}

\usepackage{pifont}

\usepackage{tabularx}
\usepackage{diagbox}
\usepackage{arydshln}
\usepackage{makecell}

\renewcommand\footnotetextcopyrightpermission[1]{}
\begin{document}
\pagestyle{empty}
\title{Unveiling the Potential of  Spike Streams for Foreground Occlusion Removal from Densely Continuous Views}

\author{Jiyuan Zhang}
\affiliation{%
  \institution{Peking University}
  \city{Beijing}
  \country{China}}
\email{jyzhang@stu.pku.edu.cn}

\author{Shiyan Chen}
\affiliation{%
  \institution{Peking University}
  \city{Beijing}
  \country{China}}
\email{2001212818@stu.pku.edu.cn}

\author{Yajing Zheng}
\affiliation{%
  \institution{Peking University}
  \city{Beijing}
  \country{China}}
\email{yj.zheng@pku.edu.cn}

\author{Zhaofei Yu}
\affiliation{%
  \institution{Peking University}
  \city{Beijing}
  \country{China}}
\email{yuzf12@pku.edu.cn}

\author{Tiejun Huang}
\affiliation{%
  \institution{Peking University}
  \city{Beijing}
  \country{China}}
\email{tjhuang@pku.edu.cn}

\renewcommand{\shortauthors}{Zhang, et al.}

\begin{abstract}
The extraction of a clean background image by removing foreground occlusion holds immense practical significance, but it also presents several challenges. Presently, the majority of de-occlusion research focuses on addressing this issue through the extraction and synthesis of discrete images from calibrated camera arrays. Nonetheless, the restoration quality tends to suffer when faced with dense occlusions or high-speed motions due to limited perspectives and motion blur. To successfully remove dense foreground occlusion, an effective multi-view visual information integration approach is required. Introducing the spike camera as a novel type of neuromorphic sensor offers promising capabilities with its ultra-high temporal resolution and high dynamic range. In this paper, we propose an innovative solution for tackling the de-occlusion problem through continuous multi-view imaging using only one spike camera without any prior knowledge of camera intrinsic parameters and camera poses. By rapidly moving the spike camera, we continually capture the dense stream of spikes from the occluded scene. To process the spikes, we build a novel model \textbf{SpkOccNet}, in which we integrate information of spikes from continuous viewpoints within multi-windows, and propose a novel cross-view mutual attention mechanism for effective fusion and refinement. In addition, we contribute the first real-world spike-based dataset \textbf{S-OCC} for occlusion removal. The experimental results demonstrate that our proposed model efficiently removes dense occlusions in diverse scenes while exhibiting strong generalization.
\end{abstract}


\maketitle

\begin{figure}[t]
    \begin{center}
        \includegraphics[width=1.0\linewidth,trim={0 0 0  0},clip]{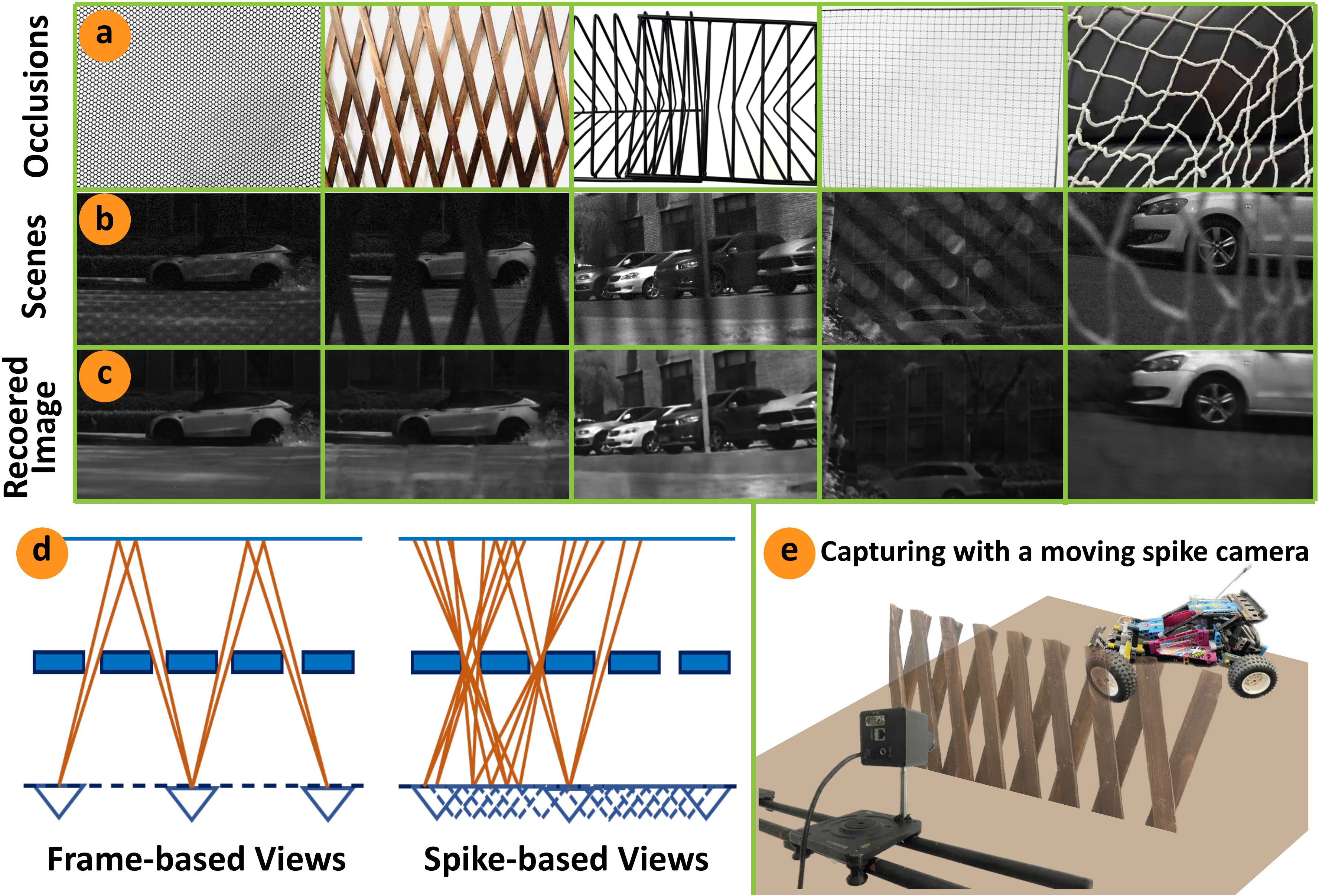}    
    \end{center}
    
    \caption{Illustration of the idea of exploring occlusion removal with a single spike camera. (a) Selected challenging occlusions, (b) Real-world Scenes with occlusions, (c) Recovered background images with the proposed \textbf{SpkOccNet}, (d) Difference between frames and spikes, (e) How we capture dataset with the spike camera.}
    \label{fig_head}
\end{figure}

\section{Introduction}
The presence of dense occlusions poses challenges to visual algorithms. Recently, frame-based algorithms have been proposed to remove such occlusions~\cite{zhang2017synthetic,wang2020deoccnet,zhang2021removing,li2021mask4d,zhang2022light,hur2023see}. However, these algorithms often rely on discrete multi-view exposures, which may not provide sufficient background information. Additionally, obtaining sharp frames in high-speed scenarios is challenging. In applications like autonomous driving, effectively removing foreground occlusions (e.g., fences) improves the perception of the environment, especially at high driving speeds. Therefore, acquiring continuous and sharp multi-view information in high-speed scenarios remains a significant challenge.

Recent years, neuromorphic sensors~\cite{chen2011pulse,dong2019efficient,gallego2020event,dong2021spike,brandli2014240,liu2019event,huang2017dynamic,9138762,huang221000x} have shown remarkable performance in various visual tasks. Typically, neuromorphic sensors generate continuous signals in an asynchronous manner, enabling its high temporal resolution sampling. Currently, two common types of neuromorphic sensors are event cameras and spike cameras. Event cameras ~\cite{brandli2014240,liu2019event,huang2017dynamic,9138762,lichtsteiner2008128} asynchronously fire events in a differential manner when the change of light intensity surpass a threshold, thus capturing rich motion information. A few studies have utilized event cameras to aid in dense occlusion removal tasks~\cite{zhang2021event,yu2022learning,liao2022synthetic}. However, event-based algorithms often require refocusing events to align them and provide more accurate information for background reconstruction, which relies on precise camera intrinsic parameters and distance information between objects and the camera, limiting its applicability. 
On the other hand, spike cameras~\cite{dong2019efficient,dong2021spike} mimics the sampling mechanism of the fovea in the retina~\cite{masland2012neuronal,wassle2004parallel}, with each pixel continuously capturing photons and firing spikes asynchronously when the accumulated intensity surpass a dispatch threshold. The integration mechanism of spike streams enables recording of absolute light intensity~\cite{huang221000x,zheng2021high,zhu2019retina}, offering more texture information for reconstructing occluded regions. Also, as shown in Fig.~\ref{fig_head}(d), the spike stream output by a spike camera provides more continuous and dense motion cues for foreground occlusion removal than conventional frame-based cameras. In this paper, we propose, for the first time, to utilize spike cameras for foreground occlusion removal tasks, demonstrating the potential of spike cameras in removing occlusions and reconstructing sharp backgrounds.

\textbf{How do we define the spike-based occlusion removal task?} 
Conventional cameras often suffer from motion blur when capturing moving scenes, limiting the acquisition of multiple perspective views with a single camera. Frame-based algorithms rely on camera arrays to compensate for the limited viewpoints, restricting their applicability in real-world scenarios. In contrast, event-based algorithms leverage a single event camera to obtain continuous imaging from different perspectives. However, these methods often rely on prior knowledge, such as camera poses, to establish the light field. Our goal is to achieve foreground removal using only one spike camera without complex equipment or calibration. Therefore, the spike-based occlusion removal task we defined possesses the following advantages: Due to the high temporal resolution of the spike camera, we are not constrained by the motion speed of the scene, and a single spike camera is sufficient to capture different continuous views. Moreover, our proposed algorithm utilizes rich information from continuous viewpoints without manual calibration, providing ample texture information for occlusion removal.

\textbf{How do we design the model?}
To deal with the spike streams, we build an end-to-end model named \textbf{SpkOccNet} for the task. Specifically, to exploit the rich temporal and spatial information in the spike stream,  we propose the Multi-View Multi-Window (MVMW) module, which integrates pulses from different time windows, each of which contains dense view information. On one hand, we choose longer windows for the spike streams, which introduces blurring effects on fast-moving foreground objects. Since the background is farther and experiences smaller relative displacement, it is less affected by blurring and provides accurate texture information with longer exposure windows.
On the other hand, for spike streams further away from the timestamp to be reconstructed, we use shorter time windows. As occluding objects exhibit greater motion relative to the central position in these windows, they contain the originally occluded regions at the reconstruction moment, serving as auxiliary information for reconstructing the central scene.
Furthermore, to handle the features extracted from dense views, we propose the Mutual Spatial-Channel Attention (MSCA) module that offers cross-view information with spatial and channel attention mechanisms. 

To improve the generalization of our algorithm in real-world scenarios, we constructed the first real spike-based occlusion removal dataset \textbf{S-OCC}. As depicted in Fig.\ref{fig_head}(e), we mounted a spike camera on a slider and moved it rapidly to capture various outdoor scenes with different occlusions. Fig.\ref{fig_head}(a) illustrates the occlusion types used, Fig.\ref{fig_head}(b) shows the occluded scenes in S-OCC, and Fig.\ref{fig_head}(c) displays the de-occlusion results obtained using the proposed \textbf{SpkOccNet}.

The contributions of this paper are summarized as follows:
\begin{itemize}
\item We pioneer the utilization of continuous viewpoint information from spike streams for foreground occlusion removal. Our approach incorporates information from different viewpoints and window lengths, leveraging mutual attention to fuse and refine the data. This eliminates the need for additional priors and enhances reconstruction results.

\item We contribute the first real-world spike-based dataset for occlusion removal, improving the algorithm's generalization in real-world scenes.

\item Experiments demonstrate the effectiveness of our algorithm in achieving promising occlusion removal results. Notably, our approach relies solely on a single camera without requiring any camera intrinsic parameters.
\end{itemize}


\begin{figure}[t]
    \begin{center}
        \includegraphics[width=1.0\linewidth,trim={0 0 0  0},clip]{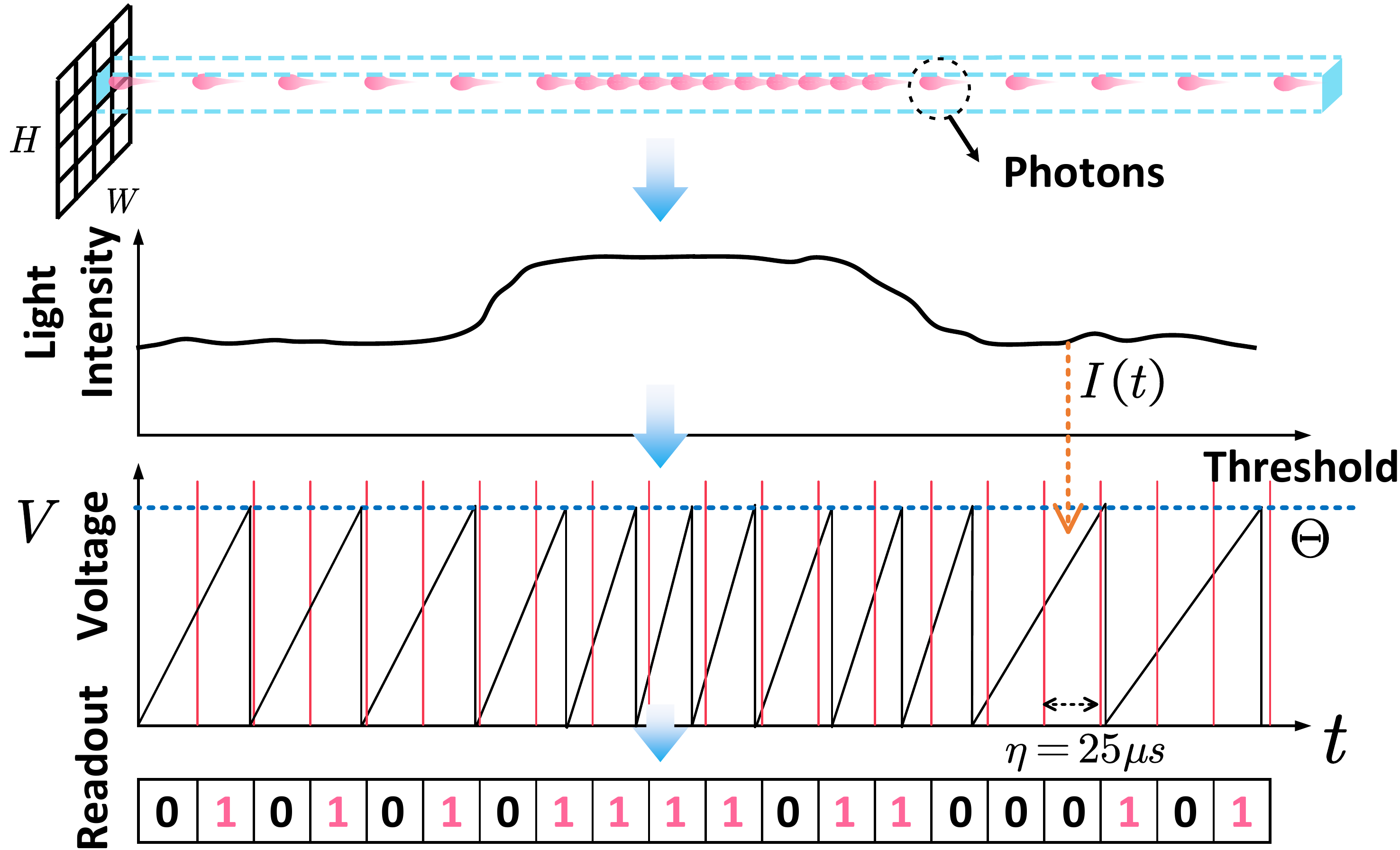}    
    \end{center}
    
    \caption{Illustration of how the spike camera firing spikes when light intensity changing.}
    \label{fig_vidar}
\end{figure}

\begin{figure*}[t]
    \begin{center}
        \includegraphics[width=0.90\linewidth,trim={0 0 0  0},clip]{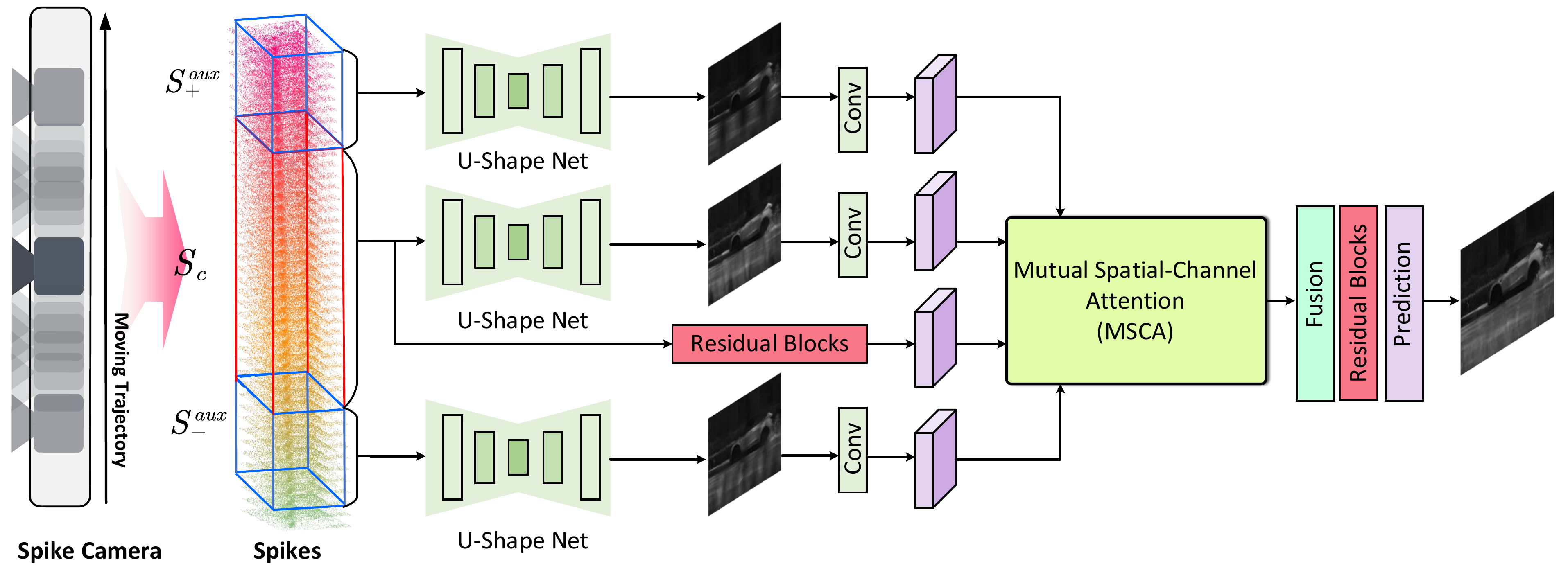}    
    \end{center}
    
    \caption{Architecture of the proposed SpkOccNet. With the spike camera moving, the continuous spike stream $S$ is fed into the network, processed by the Multi-View Multi-Window Module (MVMW) firstly then Mutual Spatial-Channel Attention (MSCA).}
    \label{fig_arch}
\end{figure*}

\section{Related Works}
\subsection{Spike-based Image Reconstruction}
Spike cameras possess several advantages, including high temporal resolution, high dynamic range, and rich preservation of spatial texture. These advantages have led to wide applications in various downstream tasks, such as optical flow estimation~\cite{hu2022optical,zhao2022learning}, object tracking~\cite{9985998}, and depth estimation~\cite{zhang2022spike,wang2022learning}. Among these tasks, the reconstruction task serves as the fundamental basis. In the early stages, Zhu et al.~\cite{zhu2019retina} propose to approximate the light intensity by statistically analyzing the spike stream. Zhu et al.~\cite{zhu2020retina,zhu2021neuspike} and Zheng et al.~\cite{zheng2021high} also develop biologically inspired reconstruction algorithms. Recently, Zhao et al.~\cite{zhao2021spk2imgnet} recover high-quality reconstructed images from spike streams using a supervised approach. Chen et al.~\cite{chen2022self} further explore self-supervised reconstruction methods. Moreover, some studies~\cite{xiang2021learning,zhao2021super} focus on recovering super-resolution images from spike stream.
However, these approaches cannot be directly applied to occlusion removal tasks due to their assumption of spike streams and reconstructed images belonging to the same scene. Existing methods also tend to use only a small segment of the spike stream as input, failing to fully exploit the spike camera's rich temporal resolution.

\subsection{Synthetic Aperture Imaging}

\textbf{Image-Based Algorigthms} Synthetic aperture imaging is a feasible route that aims to see the background scenes through occlusions assisted by multi-view images. An earlier work~\cite{vaish2004using} utilized a camera array to align the information from multiple viewpoints to a reference viewpoint using coordinate relationships. However, the planar camera array in this approach requires stringent hardware calibrations. Vaish et al.~\cite{vaish2006reconstructing} take medians and entropy into consideration and proposed a more robust cost function. Zhao et al.~\cite{pei2013synthetic} formulate an energy minimization problem to recognize each pixel from various views whether belongs to the occlusion. Zhang et al.~\cite{zhang2017synthetic} utilize a moving camera with its IMU data as the clue. Later method~\cite{yang2014all} is capable of predicting all-in-focus images. DeOccNet~\cite{wang2020deoccnet} includes a residual atrous spatial pyramid pooling module to enlarge receptive fields. Zhang et al.~\cite{zhang2021removing} use shifted micro-lens images with a dynamic filter to explore information in the light field. Recent works mainly utilize stronger CNNs to remove occlusions~\cite{li2021mask4d,zhang2022light,hur2023see}.

\noindent\textbf{Event-Based Algorigthms}
However, discrete images  captured with traditional cameras fail to provide sufficient information in scenarios with extremely dense occlusions due to their limited viewpoints. Event cameras have shown their potential on seeing through dense occlusions~\cite{zhang2021event,yu2022learning,liao2022synthetic} due to their high temporal resolution. Zhang et al.~\cite{zhang2021event} propose a hybrid network with SNNs as the encoder and CNNs as the decoder. However, the event-refocusing process in this work requires the camera intrinsic. the translation matrices of the camera and the target depth prior, which are complicated settings. Later work~\cite{liao2022synthetic} combines events and images to improve performance.

\section{Spike Generation Mechanisms}
Event cameras sample the scene radiation differentially, generating events when the change of light intensity reaches a certain threshold. In contrast, spike cameras employ an integrative sampling approach. The photosensitive units of a spike camera consist of an array of $H\times W$ pixels, with each pixel independently capturing photons continuously. The photoelectric conversion unit transforms the captured photons into electrical current $I_{x,y}(t)$ and accumulates voltage $V_{x,y}$. When the voltage $V_{x,y}$ exceeds a dispatch threshold $\Theta$, the pixel fire a spike, and subsequently the voltage $V_{x,y}$ is reset to zero. The entire process can be represented by the following equation:
\begin{equation}
    V_{x,y}^{+}(t) = \left\{
    \begin{aligned}
         & V_{x,y}^{-}(t) + I_{x,y}(t), & \text{if} ~ V_{x,y}^{-}(t) < \Theta, \\
         & 0, & \text{otherwise,}\\
    \end{aligned}
    \right.
\end{equation}
where $V_{x,y}^{-}(t)$ and $V_{x,y}^{+}(t)$ denotes the voltage before and after receiving the electric current $I_{x,y}(t)$. Since the threshold arrival time can be arbitrary, while the voltage is readout only at discrete moments within a 25$\mu$s period, the spike camera ultimately outputs a spike stream in an $H\times W \times T$ matrix.


\begin{figure*}[t]
    \begin{center}
        \includegraphics[width=0.92\linewidth,trim={0 0 0  0},clip]{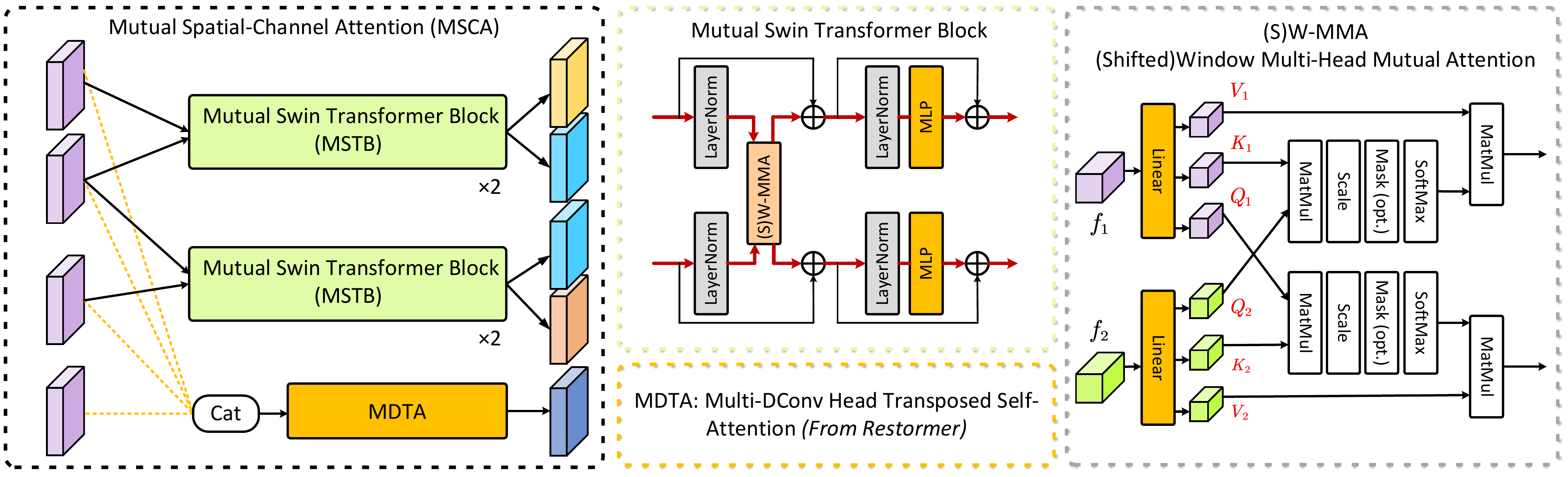}    
    \end{center}
    
    \caption{Illustration of the structure of the proposed Mutual Spatial-Channel Attention (MSCA) module, consisting of two Mutual Swin Transformer Blocks (MSTB) and a Multi-DConv Head Transporsed Self-Attention (MDTA). The MSTB is proposed two to deal with features from two branches. we build it inspired by Swin Transformer~\cite{liu2021swin} to  reduce computational complexity. The MDTA is  adapted from the Restormer~\cite{zamir2022restormer}.}
    \label{fig_msca}
\end{figure*}

\section{Methods}
\subsection{New Spike-based Dataset: S-OCC}
Occlusion removal with spike cameras is a previously unexplored area, lacking relevant existing datasets to this work. Thus, we are dedicated to constructing the first dataset based on spike cameras with various occlusions and ground truths. We name the new dataset as \textbf{S-OCC}.

\noindent\textit{\textbf{How We Set the Camera.}} In contrast to event-based and image-based approaches, this work strives for independence from camera intrinsic and extrinsic parameters, and scene prior knowledge, relying on a single spike camera. To record the scene, we mounted a spike camera on a slider and orchestrated rapid movements during each capture, with a moving time of approximately 0.1s.

\noindent\textit{\textbf{What the Occlusions are.}} We meticulously set five intricate occlusions to both enhance model robustness and unveil the potential of spike cameras. They encompassed \textbf{(1)} \textit{a square iron mesh}, \textbf{(2)} \textit{a hexagonal iron mesh}, \textbf{(3)} \textit{a dense iron frame}, \textbf{(4)} \textit{a fence}, and \textbf{(5)} \textit{an irregular fabric net}. As Fig.~\ref{fig_head}(a) visually depicted, these occlusions characterize diverse levels of sparsity and density, thereby introducing complexity to this work. Notably, occluding objects were allowed their own motion during the camera capturing process.

\noindent\textit{\textbf{How We Construct the Dataset.}} We recorded various outdoor scenes utilizing the aforementioned camera motion and occlusion setups, yielding a total of 128 sequences. Among these, 108 sequences were randomly picked for training, while the remaining 20 sequences are for testing. For static scenes without occlusion, We fixed the spike camera to capture the background scenes and  obtained grayscale images by calculating the spike firing rate~\cite{zhu2019retina}, which served as the ground truth for the dataset. As a result, each sample in this dataset comprises a spike stream alongside a clear background image with no occlusions.

\subsection{Overall Architecture}
\label{sec42}
We aim to predict the background image $I$ without occlusions through continuous spikes. We build the model called \textbf{SpkOccNet}, as shown in Fig.~\ref{fig_arch}. Due to the high temporal resolution of the spike camera (sampling frequency of 20,000Hz), each input sample is a spike stream denoted as $S$, generated by the camera through rapid sliding motion. $S$ is in size of $H \times W \times T$, where $H \times W = 250 \times 400$ represents the spatial resolution and $T=2100$ corresponds to the length of spikes (the time interval of 105ms). The spike stream $S$ is divided into three pieces for our architecture:
\begin{align}
    S = S_-^{aux} + S_c + S_+^{aux},
\end{align}
\noindent where $S_-^{aux} = H \times W \times T_-$, $S_c = H \times W \times T_c$, $S_+^{aux} = H \times W \times T_+$, and $T_-=T[0:W_{aux}], T_c=T[W_{aux}:-W_{aux}],T_+ = T[-W_{aux}:]$, as shown in Fig.\ref{fig_arch}. Note that $W_{aux}$ is a hyper-parameter that controls the length of windows. We set $W_{aux}$ to 300. For $S_c$, a longer time window can effectively blur the foreground occlusions and provide an approximate texture structure of the background. The $S_-$ and $S_+$ adjacent to $S_c$ maintain shorter time windows to serve as auxiliary inputs, providing clear texture references for recovering the background at the central moment. 

The SpkOccNet includes two stages, one is Multi-View Multi-Window Module (MVMW) and the other is Mutual Spatial-Channel Attention (MSCA). In the first stage, each input is processed by a U-shape network comprised of encoders and decoders, respectively. We denote them as $M^{enc-de}_{c}$, $M^{enc-dec}_{-}$, and $M^{enc-dec}_{+}$, and they have same structures. The output of three branches is the predicted grayscale images $\{\hat{I}\}_{c,-,+}$, denoted as $\hat{I}_c$, $\hat{I}_-$, and $\hat{I}_+$. 

In the second stage, a residual convolutional block is utilized to extract features $f_{res}$ with the original resolution, and $\{\hat{I}\}_{c,-,+}$ are processed with conv layers to obtain refined shallow features $f_c^{ref}$, $f_-^{ref}$ and $f_+^{ref}$. Then, these features from different branches are input to MSCA module $M_{\text{MSCA}}$ for feature fusion with the proposed attention mechanism in which features are enhanced with attention from others along both spatial and channel dimensions. The MSCA is formulated as followings:
\begin{align}
    \textbf{F}_{attn} = f_{M_{\text{MSCA}}} ( [f_c^{ref}, f_-^{ref}, f_+^{ref}, f_{res}] : \theta_{M_{\text{MSCA}}}),
\end{align}

\noindent where $\textbf{F}_{attn}$ is a feature list after attention-based feature aggregation and $\theta$ denotes parameters for optimization. Details of the MSCA are described in Sec.\ref{sec44}. At the end of the SpkOccNet, one conv fusion layer $M_{fuse}$, one residual block $M_{res}$ and one conv prediction layer $M_{pred}$ are sequentially arranged to further decode $\textbf{F}_{attn}$ and outputs the final predicting image $\hat{I}$, formulated as:
\begin{align}
    \hat{I} = f_{M_{pred},M_{res},M_{fuse}} ( \textbf{F}_{attn} : \theta_{M_{pred},M_{res},M_{fuse}}),
\end{align}

\begin{figure*}[t]
    \begin{center}
        \includegraphics[width=0.92\linewidth,trim={0 0 0  0},clip]{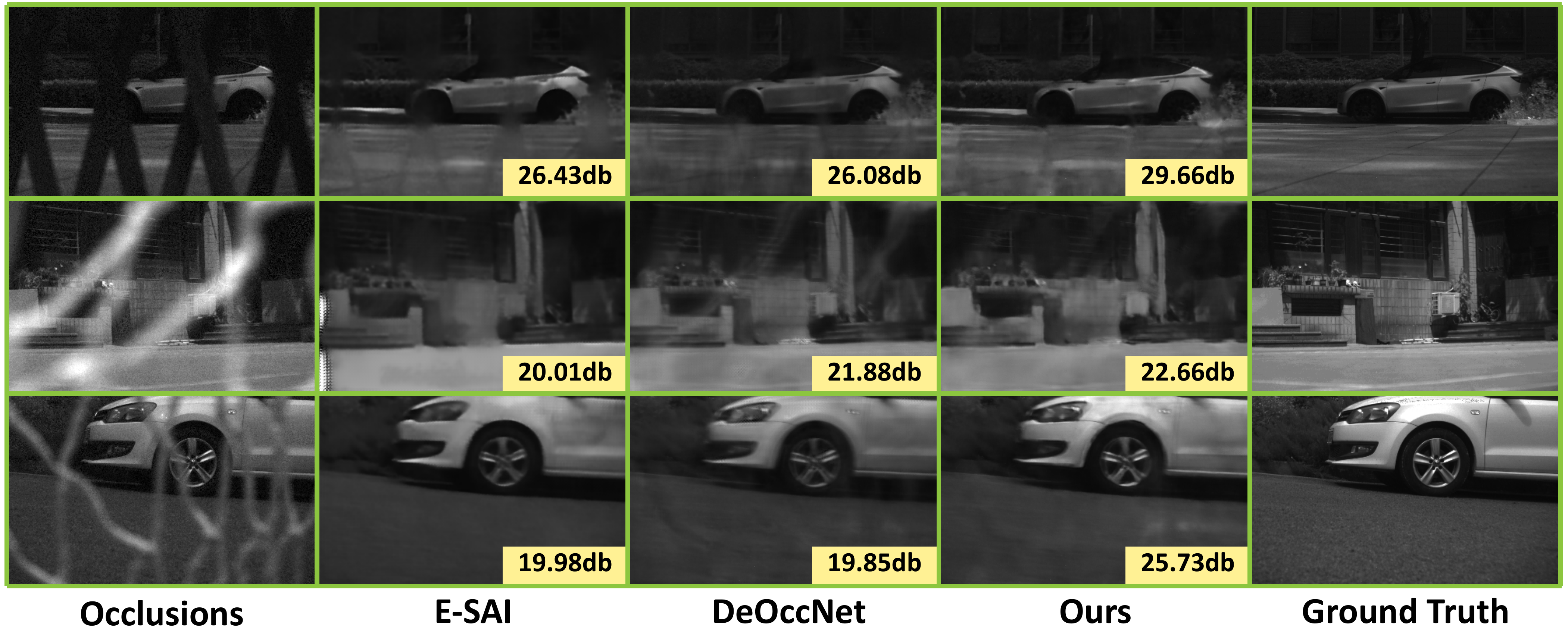}    
    \end{center}
    
    \caption{Visualized results on S-OCC dataset for comparison with other methods.}
    \label{fig_res}
\end{figure*}
\subsection{Multi-View Multi-Window Module}
\label{sec43}
As in Sec.~\ref{sec42}, we split the input $S$ into three windows, $S_-^{aux}$, $S_c$, $S_+^{aux}$, each of which contains continuous views. The timestamp for reconstruction is the center of $S_c$. Thus, $S_c$ provides spatial texture structure for reconstructing the background. During the process of capturing with the moving spike camera, the depth difference between foreground occlusions and the background is significant. Therefore the foreground occlusions will exhibit larger displacements on the imaging plane than the background. For spikes at different moments, the occluded parts in the background will change accordingly due to the variations in occlusions. Therefore, we set the longer window for $S_c$ and accumulate spikes in $S_c$ along the time axis. This transforms the foreground occlusions into a blurred effect similar to a long-exposure image, allowing the occluded background texture to have the "partially-see-through effect".

However, the long time window of $S_c$ introduces a certain level of blurring of the background. To address the issue, we utilize two shorter windows, $S_-$ and $S_+$, which are farther away from $S_c$, as auxiliary inputs. $S_-$ and $S_+$ possess complementary information to $S_c$ for two reasons: \textbf{(a)} Their foreground occlusions will exhibit larger motion displacements, exposing the occluded parts in $S_c$; \textbf{(b)} The shorter time window can provide clearer texture information.

In this stage, three U-shape network takes three windows of spikes as input and predicts the ground truth image $I$, independently. Due to the different window lengths and different occluded regions caused by motion displacement in these three inputs, the three reconstructed images $\{\hat{I}\}_{c,-,+}$ are expected to possess complementary information, meaning that the local parts with high reconstruction quality may vary among them. These three predicted images are calculated $L_1$ loss with the $I$, respectively, to provide supervised information for $M^{enc-de}_{c}$, $M^{enc-dec}_{-}$, and $M^{enc-dec}_{+}$.

\subsection{Mutual Spatial-Channel Attention}
\label{sec44}
As described above, three outputs in the first stage of SpkOccNet are complementary. Therefore, in the second stage, we propose a novel Mutual Spatial-Channel Attention (MSCA) module to fuse and refine the features from the first stage. 

In this stage, the MSCA takes four features as input, $f_c^{ref}$, $f_-^{ref}$, 
$f_+^{ref}$ which are derived from $\{\hat{I}\}_{c,-,+}$ through a conv layer, and $f_{res}$ which are derived from $S_c$ through a residual block. The residual block consists of two conv+relu layers and a skip connection from input to output, utilized to extract shallow features from $S_c$. 

As shown in Fig.~\ref{fig_msca}, MSCA includes two modules with different attention mechanisms, one is the Multi-DConv Head Transposed Self-Attention (MDTA) $M_{\text{MDTA}}$that is the same as in Restormer~\cite{zamir2022restormer}, the other is our proposed mutual swin transformer block (MSTB) $M_{\text{MSTB}}$ which is inspired by Swin Transformer~\cite{liu2021swin}. 

Before MDTA, features are concatenated together along the channel axis. The structure of MDTA, totally the same as what in Restormer, yields multi-head self-attention across channels and outputs a refined feature $f_{\text{MDTA}}$. We use the MDTA to extract valid channels across feature maps from different branches. It can be formulated as 
\begin{align}
    f_{\text{MDTA}} = f_{M_{\text{MDTA}}} ( \text{Concat}(f_c^{ref}, f_-^{ref}, f_+^{ref}, f_{res}) : \theta_{M_{\text{MDTA}}}),
\end{align}

\noindent Among $f_c^{ref}$, $f_-^{ref}$ and $f_+^{ref}$, $f_c^{ref}$ represents the center long-window textures of $S_c$ while $f_-^{ref}$ and $f_+^{ref}$ offering textures from the shorter time windows on both sides from $S_c$. We consider using the cross-view mutual attention mechanism for the following reasons:

\noindent \textbf{(A)} The occluded regions in $f_-^{ref}$ and $f_+^{ref}$, $f_c^{ref}$ differ due to the differences in viewpoints. Mutual attention can help to compensate for the occluded parts in one feature map by referring to the non-occluded parts in other feature maps. 

\noindent \textbf{(B)} Due to the high-speed camera motion, there can still be some spatial displacement of the background scenes. The mutual attention mechanism can effectively align features to mitigate the impact of camera motion.

The MSTB approximately comprises two Swin Transformer blocks~\cite{liu2021swin}, but we modify the (Shifted) Window Based Multi-Head Self Attention (\textit{(S)W-MSA}) into the (Shifted) Window Multi-Head Mutual Attention (\textit{(S)W-MMA}). Specifically, the original W-MSA takes one input $f$, obtains its \textbf{$Q$}(\textit{query}), \textbf{$K$}(\textit{key}), and \textbf{$V$}(\textit{value}) matrices and operates self attention. For our W-MMA, it takes two inputs $f_1$ and $f_2$, obtains \textbf{$Q_1$}, \textbf{$K_1$}, \textbf{$V_1$} and \textbf{$Q_2$}, \textbf{$K_2$}, \textbf{$V_2$} matrices, and operates mutual attentions as followings:
\begin{align}
    f_{1,2}, f_{2,1} &= \textbf{MutualAttention}(f_1, f_2), \\
    &= \textbf{Attention}(Q_1, K_2, V_2), \textbf{Attention}(Q_2, K_1, V_1), \\
    &= \text{softmax}(\frac{Q_1 K_2^T}{\sqrt{d_{k1}}}) V_2, \text{softmax}(\frac{Q_2 K_1^T}{\sqrt{d_{k2}}}) V_1,
\end{align}

\noindent where $\frac{1}{\sqrt{d_{k1}}}$, $\frac{1}{\sqrt{d_{k2}}}$ denotes the scaling factor. In the MSTB, we compute mutual attention twice, one between $f_c^{ref}$ and $f_-^{ref}$, the other between $f_c^{ref}$ and $f_+^{ref}$. The process can be formulated as:
\begin{align}
    f_{c,-}, f_{-,c} &= \textbf{MutualAttention}(f_c^{ref}, f_-^{ref}),\\
    f_{c,+}, f_{+,c} &= \textbf{MutualAttention}(f_c^{ref}, f_+^{ref}).
\end{align}

\noindent Note that here we omit other details in \textbf{Attention} as shown in the Fig.~\ref{fig_msca}. In this way, the MSCA stage outputs a feature list $\textbf{F}_{attn} = [f_{c,-}, f_{-,c}, f_{c,+}, f_{+,c}, f_{\text{MDTA}}]$.

\subsection{Loss Functions}
Given the ground truth image $I$, we operate $L_1$ loss with $\hat{I}_-$, $\hat{I}_+$, $\hat{I}_c$ from the first stage, and $\hat{I}$ from the second stage, which is formulated as followings:
\begin{align}
    \mathcal{L}  = \Vert  I -  \hat{I} \Vert_1 + \lambda(\Vert  I -  \hat{I}_- \Vert_1 + \Vert  I -  \hat{I}_+ \Vert_1 + \Vert  I - \hat{I}_c \Vert_1),
\end{align}

\noindent where $\lambda$ is a hyper-parameter that controls the weight of losses. ($\lambda$ set to 0.4 in later experiments.)

\section{Experiments}
\subsection{Implementation Details}
We train all networks with PyTorch with a batch size of 8 on 2 NVIDIA RTX4090 GPUs. Random cropping, random flipping, and random rotation are adopted for data augmentation. AdamW is used for optimization with an initial learning rate of 2e-4. We train networks for 2e4 iterations with the cosine scheduler. The quantitative metrics are peak signal-to-noise ratio (PSNR) and structural similarity (SSIM). In the SpkOccNet, the MSTB contains 2 blocks and each block contains 6 attention heads.

\subsection{Quantitative and Qualitative Results}
\begin{table}[!t]
\caption{ \textbf{Comparison of various de-occlusion methods on S-OCC.} }
\vspace{-2mm}
\centering
\small
\resizebox{1.0\linewidth}{!}{
\setlength{\tabcolsep}{10pt}
\renewcommand\arraystretch{1.0}
\begin{tabular}{cccc}
\bottomrule[0.15em]
\textbf{Occlusions} & \textbf{Methods} & \textbf{PSNR $\uparrow$}  & \textbf{SSIM $\uparrow$}  \\ \hline
 \hline
 \multirow{4}{*}{\textit{Fense}} & Baseline U-Net   & 19.55 & 0.782\\
                                & DeOccNet~\cite{wang2020deoccnet}  & 26.03 & 0.746\\
                                & E-SAI~\cite{zhang2021event}  & 26.68 & 0.767\\
                                & \textbf{SpkOccNet (Ours)}  & \textbf{27.36} & \textbf{0.802} \\ \hline
 \multirow{4}{*}{\textit{Raster}} & Baseline U-Net   & 23.38 & 0.704\\
                                & DeOccNet~\cite{wang2020deoccnet}  & 24.54 & 0.720\\
                                & E-SAI~\cite{zhang2021event}  & 25.13 & 0.745\\
                                & \textbf{SpkOccNet (Ours)}  & \textbf{25.45} & 0.737 \\ \hline
 \multirow{4}{*}{\textit{Square Mesh}} & Baseline U-Net   & 26.88 & 0.806\\
                                & DeOccNet~\cite{wang2020deoccnet}  & 25.89 & 0.755\\
                                & E-SAI~\cite{zhang2021event}  & 28.26 & 0.772\\
                                & \textbf{SpkOccNet (Ours)}  & \textbf{28.34} & \textbf{0.817} \\ \hline
 \multirow{4}{*}{\textit{Hexagonal Mesh}} & Baseline U-Net   & 28.44 & 0.765\\
                                & DeOccNet~\cite{wang2020deoccnet}  & 28.22 & 0.797\\
                                & E-SAI~\cite{zhang2021event}  & 27.48 & 0.776\\
                                & \textbf{SpkOccNet (Ours)}  & \textbf{29.23} & 0.756 \\ \hline
 \multirow{4}{*}{\textit{Fabric Net}} & Baseline U-Net   & 19.55 & 0.665\\
                                & DeOccNet~\cite{wang2020deoccnet}  & 20.62 & 0.664\\
                                & E-SAI~\cite{zhang2021event}  & 19.69  & 0.626\\
                                & \textbf{SpkOccNet (Ours)}  & \textbf{22.01} & 0.627 \\ \hline
 \multirow{4}{*}{\textit{Total}} & Baseline U-Net   & 24.60 & 0.772\\
                                & DeOccNet~\cite{wang2020deoccnet}  & 25.07 & 0.775\\
                                & E-SAI~\cite{zhang2021event}  & 25.51 & 0.765\\
                                & \textbf{SpkOccNet (Ours)}  & \textbf{26.46} & \textbf{0.793} \\ \hline
\end{tabular}
}
\label{table:exp}
\end{table}

In this section, we compare the proposed SpkOccNet with three other methods. Firstly, we use the U-shaped encoder-decoder network used in our model as the baseline model and train it with the same inputs. Secondly, we include DeOccNet~\cite{wang2020deoccnet}, a model that performs well in the image domain. Thirdly, we consider the model E-SAI~\cite{zhang2021event}, which is based on event cameras and trained with a hybrid SNN-CNN model. All models are trained on the proposed S-OCC dataset and evaluated on the test set using qualitative and quantitative metrics. While the Baseline U-Net, DeOccNet, and our SpkOccNet use the same inputs, we split the spike $S$ into 30 dense windows as input according to the settings in E-SAI.

The quantitative results in terms of PSNR and SSIM are presented in Tab.~\ref{table:exp}. The table shows the performance of each model on the test set for five different occlusion scenarios, as well as the average performance across all samples. It can be observed that our model, SpkOccNet, achieves the best performance on the S-OCC dataset compared to the other three methods. SpkOccNet achieves a PSNR of 26.46, which is approximately 1.4dB higher than DeOccNet and 0.95dB higher than the E-SAI method. It is worth noting that SpkOccNet has a parameter count of only about 5.8M, while DeOccNet and E-SAI have parameter counts of 39.04M and 18.59M, respectively, which are 6.7 times and 3.2 times more than ours. The performance and the parameter count demonstrate the stronger advantage of our proposed model for spike-based occlusion removal. Our model exhibits more pronounced advantages in dealing with `Fence' and `Fabric net' occlusions. `Fence' presents large-area occlusions, while `Fabric net' exhibits dense and irregular occlusions. Both of these occlusions bring significant challenges.

Fig.~\ref{fig_res} presents visualized results. Our method achieves higher-quality image reconstructions. Specifically, our method is able to recover clearer background textures and overcome the issue of the change of illumination caused by occlusions. Although the E-SAI can reconstruct relatively smooth images, the reconstructed images appear blurry and the effect of underexposure or overexposure exists in some regions. On the other hand, DeOccNet does not perform well in removing severe occlusions. The results demonstrate that our method outperforms the other two image-domain and event-domain methods.

\begin{table}[!t]
\caption{ \textbf{Ablation study of our proposed modules and different inputs.} }
\vspace{-2mm}
\centering
\small
\resizebox{1.0\linewidth}{!}{
\setlength{\tabcolsep}{10pt}
\renewcommand\arraystretch{1.0}
\begin{tabular}{cccc}
\bottomrule[0.15em]
\textbf{Architecture} & \textbf{Inputs} & \textbf{PSNR $\uparrow$}  & \textbf{SSIM $\uparrow$}  \\ \hline
 \hline
Single U-Net                        & $S_c$                 & 22.82 & 0.750\\
Single U-Net                        & $S_c$, $S_-$. $S_+$   & 24.60 & 0.772\\
+ MVMW                              & $S_c$, $S_-$. $S_+$   & 25.57 & 0.779\\
\textbf{+ MVMW + MSCA}  & $S_c$, $S_-$. $S_+$   & \textbf{26.46} & \textbf{0.793}\\ \hline
\end{tabular}
}
\label{table:abla}
\end{table}

\subsection{Ablation Studies}
To validate the effectiveness of the proposed modules in SpkOccNet, we conducted comprehensive ablation experiments, and the results are shown in Tab.\ref{table:abla}.  

Firstly, we conducted ablations on the input. In the case of using only one U-shaped encoder-decoder network $M^{enc-dec}$, we experimented with using $S_c$, $S_-$, $S_+$, and only $S_c$ as inputs respectively. The results demonstrated that the auxiliary inputs significantly contributed to recovering textures. Secondly, the third row of the table validates the effectiveness of the proposed MVMW module. The last two rows verify the impact of the MSCA module on performance. It can be observed that the presence of MSCA further improved the model's performance, indicating the effectiveness of our proposed multi-view mutual attention scheme.

\section{Conclusion}
In this work, we utilize dense spike streams for foreground occlusion removal for the first time, leveraging continuous viewpoint information. Our model SpkOccNet integrates information from different viewpoints and window lengths, employing mutual attention for effective fusion and refinement. We contribute the first real-world spike-based dataset S-OCC for occlusion removal. Remarkably, our algorithm achieves impressive occlusion removal results using a single camera without relying on any camera intrinsic parameters and camera pose information.

\clearpage
\bibliographystyle{ACM-Reference-Format}
\bibliography{ref}

\end{document}